\pdfoutput=1

\documentclass[11pt]{article}

\usepackage[preprint]{coling}

\usepackage{times}
\usepackage{latexsym}

\usepackage[T1]{fontenc}

\usepackage[utf8]{inputenc}

\usepackage{microtype}

\usepackage{inconsolata}

\usepackage{graphicx}

\usepackage{booktabs}
\usepackage{multirow}
\usepackage{enumitem}
\usepackage{longtable}
\usepackage{color}
\usepackage{colortbl}

\title{Making Large Language Models into World Models with Precondition and Effect Knowledge}

\author{Kaige Xie \quad Ian Yang \quad John Gunerli \quad Mark Riedl \\
School of Interactive Computing \\
Georgia Institute of Technology \\
\texttt{\{kaigexie, iyang30, hakancangunerli, riedl\}@gatech.edu} \\}

\begin{document}
\maketitle
\begin{abstract}
World models, which encapsulate the dynamics of how actions affect environments, are foundational to the functioning of intelligent agents.
In this work, we explore the potential of Large Language Models (LLMs) to operate as world models.
Although LLMs are not inherently designed to model real-world dynamics, we show that they can be induced to perform two critical world model functions: determining the applicability of an action based on a given world state, and predicting the resulting world state upon action execution.
This is achieved by fine-tuning two separate LLMs---one for precondition prediction and another for effect prediction---while leveraging synthetic data generation techniques.
Through human-participant studies, we validate that the precondition and effect knowledge generated by our models aligns with human understanding of world dynamics.
We also analyze the extent to which the world model trained on our synthetic data results in an inferred state space that supports the creation of action chains, a necessary property for planning.
\end{abstract}

\section{Introduction}

\begin{figure}[t]
\centering
\includegraphics[width=\linewidth]{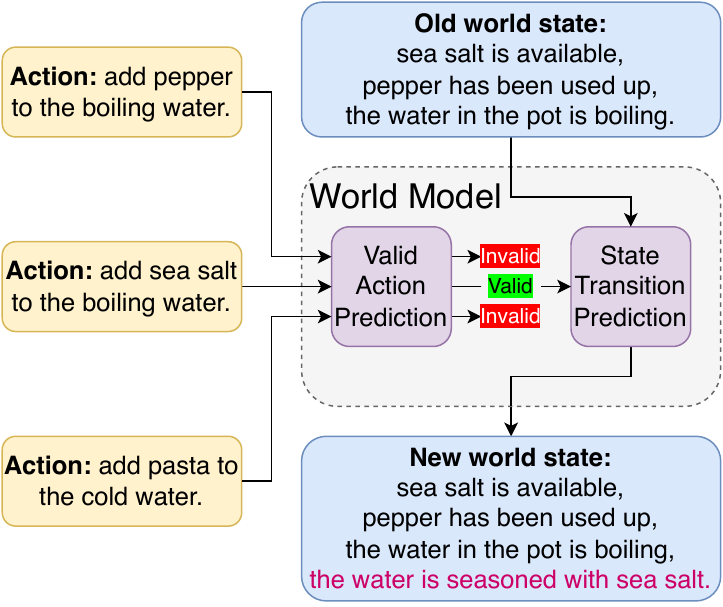}
\caption{A world model must be able to make valid action prediction and state transition prediction.
Valid action prediction refers to predicting whether an action is valid to be taken at a world state.
State transition prediction refers to predicting after a valid action is taken at an old world state, what the new world state will look like.
}
\label{fig:intro}
\end{figure}

Intelligent agents must reason about how actions affect the world.
A {\em world model} is a model of the underlying, true dynamics of how an environment can change or be changed.
A world model can also be referred to as a transition model, $P(s'|s, a)$ because it tells us how an action $a$, when performed in a state $s$, can result in the world transitioning to state $s'$. 
World models are an essential aspect of creating {\em agents}, which are entities that can perceive the environment and perform actions in order to affect change on the environment with respect to some objective. 

Are Large Language Models (LLMs) such as GPT-4~\cite{openai2023gpt4}, ChatGPT~\cite{openai2022chatgpt}, and Llama 2~\cite{touvron2023llama}, capable of expressing themselves as world models?
The answer is generally found to be ``no''~\cite{valmeekam2023investigating,wang2024can,pan2024unifying,chambersberall}.
However, some nuance is needed. 
LLMs express world model-like capabilities because they are trained on language produced by humans who operate in a world dictated by a consistent set of physical rules---water always boils when hot enough, objects always fall due to gravity, activating the brakes on a bike generally causes the bike to come to a stop, etc. 
Phenomena of the real world appear in our text corpora and are learned such that when we ask questions about the dynamics of the real world (e.g., ``what happens when I apply the brakes?''), we often get plausible responses.
However, the ability to generate a span of text that plausibly explains what happens does not necessarily mean that the LLM understands the phenomenon.
This can be demonstrated when we ask LLMs about novel situations or novel combinations of phenomenon that are too far outside their training distribution. 

Counter-intuitively, LLMs trained on sufficiently rich text sources may be induced to behave like a world model.
While an LLM may not be reasoning about the underlying transition dynamics to answer a question about the application of brakes on a bike, the same LLM can be asked questions about the applicability of actions and how the world is changed by actions carried out in the real world.
For example, large language models such as GPT-4, Gemini 1.5 Pro, and Claude 3.5 Sonnet can reliably answer questions about what must be true in the world for an action to be performable (also called {\em preconditions}), and what aspects of the world does one desire to affect through the execution of an action (also called {\em effects}).
Precondition and effect knowledge about actions can be used to assemble a world model because they tell us whether an action can be performed and what states we will transition to if an action is performed. 

In this work, we show how to induce large language models to behave like world models. 
As depicted in \autoref{fig:intro}, a world model must perform two functions.
(1)~It must determine action applicability. 
We must be able to answer the question: ``can this action be performed at this time?''
(2)~If the action is applicable, a world model must be able to answer the question: ``how the world is different if it were to be performed?''
This is equivalent to asking ``what state I will be in?''
We fine-tune two LLMs, one that predicts the preconditions of an action, and another that predicts the effects of an action. 
We provide a means for these two models to work together, along with procedures for checking preconditions against a world state and applying effects to alter the world state, thus fully replicating the functionality of a world model.
We also provide a method for using LLMs to generate synthetic data with which to fine-tune the above two models.

We conduct human-participant studies that assess the extent to which people agree that the corpus of preconditions and effects generated by our models matches their own world model understanding;
human understanding of the real world is the best source of ground-truth data available.
Then, having established bounds on the accuracy of our models with respect to human understanding, we conduct automated evaluations showing the effectiveness of our overall approach in terms of precondition/effect inference and world modeling.
We also measure the reliability of our synthetic data generation approach through human evaluations.

In summary, our contributions are as follows:
\begin{itemize}
    \item we introduce an approach to inducing an LLM to behave like a world model that is able to perform valid action prediction and state transition prediction;
    \item we design a technique to create a synthetic precondition/effect corpus from the LLM that is essential for building a world model;
    \item both human evaluations and automated evaluations show the effectiveness of our method in creating a high-quality precondition/effect corpus and a capable world model.
\end{itemize}

\section{Related Work}

\subsection{World Modeling}

World modeling has emerged in reinforcement learning as an effective way to learn both the transition function and policy for a given RL task~\cite{ha2018recurrent}.
Improvements in world modeling have led to advances in causal reasoning~\cite{yu2023explainable,richens2024robust}, world knowledge maintenance~\cite{freed2023learning,zhang-etal-2023-large,samsami2024mastering}, and model-based planning~\cite{guan2023leveraging,zhang2021world,sekar2020planning}.
For causal reasoning, \citet{yu2023explainable} explore causality in RL and find that learning a causal world model without knowledge of the environment structure can improve explainability without suffering from low accuracy.
\citet{richens2024robust} incorporate concepts from causality and decision theory to prove that agents that satisfy a large set of distribution shifts must have necessarily approximated a causal world model of the data generation process.
For world knowledge maintenance, \citet{freed2023learning} explore to what extent language models intrinsically encode relational world knowledge through the training process without a predefined relational schema.
\citet{zhang-etal-2023-large} survey different methods for incorporating world knowledge into LLMs, and distinguish between implicit methods (emergent world knowledge after training on massive corpora, knowledge editing) and explicit methods (external memory, off-the-shelf retrieval, or Internet-enhanced).
\citet{samsami2024mastering} approach the memory problem for world model-based RL agents by integrating state-space models into the DreamerV3 world model to capture and maintain long-range relational knowledge.
For model-based planning, \citet{guan2023leveraging} argue that LLMs themselves are not sufficient to tackle difficult planning problems, and propose to use pre-trained language models as an interface between planning domain definition language and natural language, so as to construct more capable explicit world domain models.
\citet{zhang2021world} propose an algorithm that first groups together goals within the world model embedded in latent space by temporal distance, then learns sparse landmarks that can be used for planning, aiming to overcome the difficulty of long planning horizons which might lead to world models that are far from reality.
\citet{sekar2020planning} demonstrate that self-supervised world models can be used to help RL agents more efficiently explore environments via planning, and better generalize to unseen tasks as well.
Recently, world models have started to have different ways of intersecting with LLMs~\cite{nottingham2023embodied,hao-etal-2023-reasoning,xiang2024language}.
\citet{nottingham2023embodied} propose to use LLMs to hypothesize a world model that will be utilized for planning \& exploration and verified with grounded experience, in order to improve sample efficiency of RL agents.
\citet{hao-etal-2023-reasoning} introduce Reasoning via Planning which leverages an LLM as a world model and applies Monte Carlo Tree Search to build a reasoning tree that is explored iteratively to provide reasoning guidance.
\citet{xiang2024language} propose a paradigm for efficiently fine-tuning LLMs with embodied experiences from world models to improve performance on seen and unseen RL tasks.
In contrast, we focus on developing techniques to induce the LLM to work reliably as a world model.

\subsection{Precondition and Effect}

Understanding action preconditions and effects in text is a crucial yet challenging task.
\citet{branavan-etal-2012-learning} pioneer work in this area using reinforcement learning to extract high-level planning knowledge from text with the guidance of action preconditions and effects.
\citet{dalvi-etal-2018-tracking} develop a dataset and models for paragraph comprehension, and highlight the importance of tracking state changes in procedural text.
\citet{hayton2020narrative} propose an automated process for extracting action models from text summaries, though their approach aims primarily to mirror the input narrative.
\citet{li2024naruto} introduce a more sophisticated system that addresses complex challenges in narrative texts such as nested event arguments and conditional events.
Their approach combines structured event extraction with predictions of commonsense event relations to get more nuanced action models.
\citet{wu-etal-2023-learning} highlight the importance of extracting pre- and post-conditions from instructional texts and introduce an approach that leverages weak supervision and contextual information to improve action condition inference.
\citet{martin2021neurosymbolic} designs a commonsense rule engine with preconditions and effects derived from VerbNet for an appropriate selection of events in neurosymbolic story generation to improve the causal consistency of generated stories.
Commonsense reasoning and its connection to preconditions have been a major area of study, particularly in relation to how AI systems infer what makes a statement valid or invalid in different contexts.
For instance, \citet{qasemi-etal-2022-pinks} design models that utilize preconditions to understand common actions and statements.
\citet{kwon-etal-2020-modeling} create a dataset to analyze how preconditions operate in textual contexts, allowing for a deeper understanding of both enabling and disabling conditions that can affect whether a given action or statement is possible.
\citet{qasemi-etal-2022-paco} formalize the notion of enabling and disabling preconditions in commonsense reasoning, and introduce a framework that forces models to make clear decisions about the preconditions required for a statement to hold true.
This differs from prior work that primarily focuses on probabilistic inferences, such as the ATOMIC dataset~\cite{sap2019atomic}, which explores cause-and-effect relationships but does not make explicit the conditions under which those relationships are valid.
\citet{kwon-etal-2021-toward} emphasize the importance of variety in precondition inference and tackle the issue of limited precondition generation by developing methods that can create a broader range of preconditions, which enhances the robustness of commonsense reasoning models.
Diversity in preconditions and effects is crucial for building AI systems that can handle the variability of real-world scenarios.
Our work aims to make accurate precondition and effect inferences by learning from a high-quality synthetic corpus of action preconditions and effects.

\section{World Model}

This section introduces our methodological contributions in creating a world model.
We focus on real-world domains where there is significant dependency among  actions, meaning that to be able to take one action in an environment, it is generally necessary to take a few other actions beforehand to set up an appropriate world state for that action to be taken.
We call this phenomenon of dependency among actions as \textit{action chaining}.
Representative real-world domains such as dish cooking with a high degree of action chaining are worth studying because it is technically more challenging to model actions and states in these domains than fictional domains existing works~\cite{ammanabrolu2021learning,ammanabrolumodeling} have been focusing on where there is much less action chaining.

A world model needs to be able to make the following two types of predictions: \textit{valid action prediction} and \textit{state transition prediction}.
Valid action prediction means: given a world state at a certain time point, predicting which actions are valid to be taken.
State transition prediction means: given a world state at a certain time point and a valid action that has been taken, predicting what the new world state is, i.e. how the world state transitions from the old one to the new one.
Previous work~\cite{ammanabrolu2021learning} tackles these two prediction tasks by directly modeling the mapping relationships between states and actions.
However, by ignoring preconditions and effects, it reduces the possibility of interpretability.
In high-stakes domains with more complicated actions and states, it is usually critical to make interpretable predictions in case developers or users might want to know how/why a certain prediction is made.
This motivates us to use a first principles approach to building the world model by modeling the preconditions and effects and then connecting them back to actions and states.
In this way, two complex prediction tasks (valid action prediction and state transition prediction) are first broken down into the most basic elements (precondition and effect of actions) and then reassembled from the ground up.

Specifically, our world model consists of two main sub-modules: a precondition/effect inference module (\S\ref{subsec:module1}) and a semantic matching module (\S\ref{subsec:module2}).
The world model performs valid action prediction and state transition prediction by invoking these two modules (\S\ref{subsec:workflow}).

\subsection{Precondition/Effect Inference Module}
\label{subsec:module1}

This module is designed to be able to take an action and infer the preconditions that this action requires, and the effects this action will cause if it is taken.
We fine-tune one LLM for precondition inference.
The input to this LLM is an action and its output is the corresponding precondition, where both the input and output are in natural language.
We similarly fine-tune another LLM for effect inference with an action as the input and effect as the output.

To train these models, we require a dataset of actions accompanied by their preconditions and effects.
However, to the best of our knowledge there is no existing precondition/effect corpus created for domains with significant action chaining.
Collecting a supervised dataset from the human annotation is an option but it might be a bit costly to carry out.
To fill this gap while saving the cost, we curate an action precondition/effect corpus automatically by prompting GPT-4~\cite{openai2023gpt4}.
Through preliminary prompting experiments, we find that GPT-4 possesses the intrinsic knowledge about action precondition/effect but does not necessarily apply that knowledge when generating plan action sequences. 
That is, a carefully designed prompting technique is needed to properly induce this knowledge out of GPT-4.

An additional challenge that needs to be overcome during the data curation is how to ensure that there is a high degree of action chaining.
From the standpoint of preconditions and effects, significant action chaining can be interpreted as follows: in most cases, one action's preconditions are covered in other actions' effects, and one action's effects are covered in other actions' preconditions as well, suggesting strong dependency among actions.

We propose a \textit{global-local} prompting technique to induce high-quality action preconditions and effects with significant action chaining.
This technique contains five steps:
\begin{enumerate}
    \item Prompt GPT-4 to come up with a full action plan which involves a series of mutually dependent action steps for performing a certain task.
    \item Prompt GPT-4 with few-shot annotated examples to selectively discard the action steps that it believes are isolated from other steps and thus do not have enough potential to produce significant action chaining, and also to optionally add new action steps if they can provide chaining and make the whole action plan more coherent.
    \item Prompt GPT-4 to simultaneously generate preconditions and effects for each action step.
    \item Prompt GPT-4 with few-shot annotated examples to (1) identify action steps whose preconditions and effects are not quite chained with other steps', (2) and next perform a one-time re-generation on the identified steps to obtain a new version of preconditions and effects, (3) and then determine whether or not to selectively discard some of these action steps due to weak action chaining.
    \item Filter action plan samples according to the percentages of the preconditions/effects that are not chained (i.e., not covered) in the plans: specifically, feed GPT-4 with both a sample's collection of preconditions and a sample's collection of effects, and next ask GPT-4 to output the preconditions/effects which are not covered in the effect/precondition collections, and then calculate the percentage number, and finally filter samples of highest percentages (we filter the highest 5\%).
\end{enumerate}
We call this prompting technique as \textit{global-local} because it first performs one round of step discard based on a local inspection of individual action steps (steps \#2 and \#4), and then performs another round of sample filtering based on a global view of the whole action plan (step \#5).
Note that since a sample technically contains a series of action steps, sample filtering can also be viewed as step discard in batch.
All the prompts we use are presented in \autoref{sec:global-local}.
We run the global-local prompting with GPT-4 thousands of times and obtain more than two thousand action plans, each of them comprising about twenty action steps on average.

\subsection{Semantic State Matching}
\label{subsec:module2}

To support the valid action prediction and state transition prediction, we need
(1) to determine whether the inferred preconditions are a subset of a world state, and
(2) to determine how the world state must change to produce a successor world state. 
Since world states are natural language descriptions, as are preconditions and effects, we require a means of semantically matching preconditions to states and updating states based on effects.

Specifically, we design two separate GPT-4-based modules for the two prediction tasks respectively.
For the valid action prediction task, its semantic matching module needs to be able to match the inferred action preconditions with the current world state, and then determine whether all the preconditions are covered in the current world state, meaning that all the preconditions are satisfied and thus this action is valid to be taken at this moment.
For the state transition prediction task, its semantic matching module needs to be able to match the inferred action effects with the current world state, and then determine whether there exists any part of the world state that contradicts the effects, meaning that this state transition requires to update the old world state by adding the effects to the old world state and in the meantime deleting the contradicted part from the old world state.
All the prompts we use are presented in \autoref{sec:semantic-matching}.

\subsection{Applying the World Model}
\label{subsec:workflow}

Recall that a world model must do two things: valid action prediction and state transition prediction.
To do valid action prediction, our world model first uses the precondition inference module to infer an action's preconditions, then uses the semantic matching module to match the inferred preconditions with the current world state, and makes a judgment on whether this action is valid to be taken at the moment.

To do state transition prediction, our world model first uses the effect inference module to infer an action's effects, then uses the semantic matching module to match the inferred effects with the current world state, and predicts the new world state by making proper addition and/or deletion to the old world state.

\section{Evaluation}

We perform multi-faceted evaluations of our world model.
Specifically, we evaluate the following three things: (\S\ref{subsec:eval-gl-prompting}) the effectiveness of the global-local prompting technique in creating a high-quality action precondition/effect corpus, (\S\ref{subsec:eval-pe-inference}) the effectiveness of the precondition/effect inference module in making accurate precondition/effect predictions, and (\S\ref{subsec:eval-world-model}) the effectiveness of the world model (precondition/effect inference module and semantic matching module working together) in making accurate valid action predictions and state transition predictions.
We choose dish cooking as the domain to evaluate as it is a representative real-world domain example in which there is significant dependency among different actions.

In addition to evaluations based on automatic metrics, we also perform human evaluations to make the assessment more comprehensive.
All the human evaluations are performed using the Prolific crowdsourcing platform\footnote{\url{https://www.prolific.co/}}.
These human evaluations have been approved by our institution's Institutional Review Board (IRB).
We qualify annotators by first asking them a screening question at the beginning of the questionnaire, and then verifying their answers manually to disregard annotations provided by those who fail the screening.
We require annotators to be physically located in the U.S. and to speak English as a first language.
For each human evaluation, we source a distinct set of annotators without any overlap to avoid potential bias in annotations that could occur from participating in related evaluations in the past.
In each evaluation, we measure the average inter-annotator agreement using Fleiss's kappa~\cite{fleiss1971measuring}.

\subsection{Evaluation of Global-Local Prompting}
\label{subsec:eval-gl-prompting}

We seek to understand whether our global-local prompting technique is effective in creating a high-quality action precondition/effect corpus that is reliable and informative enough for an inference module to learn useful knowledge from.

The reliability of the corpus is determined by how reasonable an action's preconditions and effects are.
We randomly select 30 action samples (with their preconditions and effects) from the corpus, and ask annotators to inspect each of them to judge if the preconditions and effects are reasonable for the action.
We arrange five different annotators to assess each action sample, and use the majority vote of their annotations as the final judgment.
We find that 93\% of action samples are deemed reasonable with a moderate average inter-annotator agreement, suggesting the reliability of the corpus.

The informativeness of the corpus is determined by how significant the action chaining is.
To measure it, we randomly select 30 action plan samples (each of them comprising a series of action steps with preconditions and effects) from the corpus, and ask annotators to inspect each of them to judge if there exist more than two action steps (approximately more than 10\% of all action steps in an action plan sample) whose preconditions or effects are never covered in any other action steps' effects and preconditions (i.e. indicating insignificant action chaining).
We find that 87\% of action plan samples are deemed to have significant action chaining with a moderate average inter-annotator agreement, suggesting the informativeness of the corpus.

Since this technique involves multiple steps, we need to confirm its effectiveness by evaluating if the key steps we designed are running properly as we expected.
We specifically examine steps \#2, \#4, and \#5 (global and local step discard), because steps \#1 and \#3 are foundational steps that have been implicitly measured in the evaluation of the informativeness and reliability of the corpus.

For steps \#2 and \#4 (local step discard), we measure how insignificant the action chaining is in the discarded action steps.
We randomly select 30 discarded action step samples (15 from step \#2 without preconditions/effects, and 15 from step \#4 with preconditions/effects), provide them along with their corresponding original action plans to annotators, and ask annotators to inspect each of them to judge if they are not chained with other steps in the action plans.
We find that 73\% of discarded action step samples are deemed not chained with other steps in action plans with a moderate average inter-annotator agreement, suggesting the effectiveness of steps \#2 and \#4.

For step \#5 (global step discard), we measure how insignificant the action chaining is in the filtered action plans.
We randomly select 30 filtered action plan samples, and also randomly select 30 kept action plan samples from the corpus.
We put the selected samples into 30 filtered-kept sample pairs (one filtered and one kept) and ask annotators to inspect each of them to judge if the filtered sample clearly has less action chaining than the kept sample.
We find that 80\% of the time the filtered action plan samples have less action chaining than the kept samples with a moderate average inter-annotator agreement, suggesting the effectiveness of step \#5.

\subsection{Evaluation of Precondition/Effect Inference Module}
\label{subsec:eval-pe-inference}

\begin{table}
    \small
    \resizebox{\columnwidth}{!}{
    \centering
    \begin{tabular}{lccccc}
        \cmidrule[\heavyrulewidth]{1-6}
        & F1 & BLEU-2 & BLEU-3 & ROUGE-L & SMS \\
        \cmidrule[\heavyrulewidth]{1-6}
        \rowcolor[gray]{0.95} \multicolumn{6}{c}{\it Precondition Inference} \\
        Ablation-Local & 58.57 & 63.47 & 57.88 & 51.07 & 17.02 \\
        Ablation-Global & 60.53 & 66.25 & 60.06 & 52.24 & 17.89 \\
        Full Method & \textbf{65.67} & \textbf{70.08} & \textbf{64.99} & \textbf{57.96} & \textbf{19.77} \\
        \rowcolor[gray]{0.95} \multicolumn{6}{c}{\it Effect Inference} \\
        Ablation-Local & 55.03 & 60.41 & 54.70 & 53.17 & 16.56 \\
        Ablation-Global & 58.51 & 62.71 & 57.92 & 55.13 & 17.55 \\
        Full Method & \textbf{61.43} & \textbf{65.20} & \textbf{59.35} & \textbf{57.72} & \textbf{18.25} \\
        \cmidrule[\heavyrulewidth]{1-6}
    \end{tabular}
    }
    \caption{Results on automatic metrics for precondition inference module and effect inference module.} 
    \label{tab:automatic-result-1}
\end{table}
\begin{table}
    \small
    \resizebox{\columnwidth}{!}{
    \centering
    \begin{tabular}{lcccccc}
        \cmidrule[\heavyrulewidth]{1-7}
        & Acc. & F1 & BLEU-2 & BLEU-3 & ROUGE-L & SMS \\
        \cmidrule[\heavyrulewidth]{1-7}
        Ablation-Local & 74.50 & 49.73 & 55.21 & 49.83 & 44.26 & 12.96 \\
        Ablation-Global & 77.00 & 53.43 & 59.09 & 54.30 & 49.71 & 14.52 \\
        Full Method & \textbf{81.50} & \textbf{56.05} & \textbf{61.69} & \textbf{56.18} & \textbf{52.75} & \textbf{15.19} \\
        \cmidrule[\heavyrulewidth]{1-7}
    \end{tabular}
    }
    \caption{Results on automatic metrics for world model's valid action prediction and state transition prediction.} 
    \label{tab:automatic-result-2}
\end{table}

In this evaluation, we seek to understand whether our precondition/effect inference module is effective in making accurate precondition/effect predictions.
We train two FLAN-T5 models~\cite{chung2024scaling}, \texttt{flan-t5-large}\footnote{\url{https://huggingface.co/google/flan-t5-large}}, on our action precondition and effect corpus to learn to perform precondition and effect inference separately.
Implementation details can be found in \autoref{sec:appendix-implementation-details}.
We maintain a holdout test set which is composed of 200 action plans with a series of action steps.
We use the following four automatic evaluation metrics to measure the empirical test-set performance: (1) token-level F1 score~\cite{tjong-kim-sang-de-meulder-2003-introduction} (F1); (2) BLEU score~\cite{papineni-etal-2002-bleu} (BLEU-2 and BLEU-3); (3) ROUGE score~\cite{lin-2004-rouge} (ROUGE-L); (4) sentence mover's similarity~\cite{clark-etal-2019-sentence} (SMS).

Since the empirical performance of the precondition/effect inference module is dependent on how good the training data is, we thereby perform ablation studies to investigate how the key steps (global and local step discard) in creating the corpus may impact the precondition/effect inference performance, and also to confirm the key steps' necessity by making performance comparison between the full version and the ablated versions.
Specifically, we study the following two ablations: training inference modules on the corpus created by the global-local prompting technique while skipping (1) steps \#2 and \#4 (denoted as Ablation-Local) or (2) step \#5 (denoted as Ablation-Global).
Results presented in \autoref{tab:automatic-result-1} show that the precondition/effect inference modules trained on our high-quality corpus can make accurate predictions that are very consistent with the ground truth.
The performance comparison between the full version and the ablated versions demonstrates that both global and local step discard are necessary and beneficial for creating a high-quality corpus that LLMs can further learn from to make reliable action precondition/effect inferences.

Automatic evaluation metrics are not always perfect since they only measure how aligned the prediction and the ground truth are.
Sometimes it may be the case that an automatic metric negatively prefers a prediction that a human would prefer positively, and vice versa.
Because of this potential discrepancy, we need human evaluation to assess the prediction more comprehensively.
We randomly select 30 model-predicted action preconditions and 30 model-predicted action effects.
We ask annotators to inspect each of the predicted preconditions and effects to judge if these predictions look reasonable to them and are consistent with their commonsense understanding of the real world.
We find that 77\% of the predicted preconditions and 70\% of the predicted effects match human's world model understanding with a moderate average inter-annotator agreement, confirming the effectiveness of the precondition/effect inference module from the human perspective.

\subsection{Evaluation of World Model}
\label{subsec:eval-world-model}

In this evaluation, we seek to understand whether our world model is effective in making accurate valid action predictions and state transition predictions.
Before starting the evaluation, we perform a corpus refactoring to adapt the data into appropriate input-output formats for the two prediction tasks.
Since both tasks need the ground-truth world state as part of the input, we refactor the action preconditions and effects in an entire action plan to get the ground-truth world state at each time step.
We automate this refactoring process with the assistance of GPT-4o~\cite{openai2024gpt4o}.
For each action plan, we get the initial world state by combining all the preconditions excluding those that are covered in any of the effects within the plan.
We derive new world states right after taking every single step by using the state addition and deletion at each step to perform step-by-step iterative updates on old world states.
The state addition is straightforward to get as it is just equivalent to the ground truth of the action effects.
The state deletion is obtained by iterating through each item in the old state, and asking GPT-4o to compare them to each item in the state addition to judge if there is a contradiction and thus that item in the old state needs to be deleted.

For each of the two prediction tasks, we maintain a 200-sample holdout test set respectively.
For valid action prediction, we balance the test set by equalizing the number of valid and invalid action samples, and use the prediction accuracy (Acc.) as the evaluation metric.
For state transition prediction, we follow an existing evaluation setup~\cite{ammanabrolu2021learning} by combining the state addition and deletion as the final prediction and using the metrics introduced in \autoref{subsec:eval-pe-inference} (F1, BLEU-2/3, ROUGE-L, and SMS) to measure the performance.
Results presented in \autoref{tab:automatic-result-2} show that on both tasks our world model can make accurate predictions consistent with the ground truth.
We perform similar ablation studies as \autoref{subsec:eval-pe-inference} on Ablation-Local and Ablation-Global, and the results demonstrate that both global and local step discard are necessary and beneficial for establishing a capable world model.
For state transition prediction, we also do a similar human evaluation as \autoref{subsec:eval-pe-inference} to evaluate our world model more comprehensively. On 30 randomly selected samples, we find that with a moderate average inter-annotator agreement, 63\% of the predicted state transitions are consistent with human's commonsense understanding and reasoning of the real world.

\section{Analysis of Search Space}

One of the utilities of a world model is to make logically sound plans.
We analyze the search space of our world model to determine if it could be used by a hypothetical planning system.
Specifically, we answer the following two questions.
(1) Given a never-before-seen action, how likely is it for this action to be satisfiable by the search space our world model creates?
This question assesses the likelihood that chaining will occur for any given target, and that valid plans exist for which an action would be the last step in a plan. 
An action being satisfiable by a search space means that the preconditions of this action can be met in a certain world state that can be reached by taking some actions from this search space.
In other words, in this search space it is possible to reach a certain world state after taking some actions and applying their action effects such that it now becomes valid to take this action.
(2) Given a never-before-seen action, how many different ways are there to satisfy the action in the search space? This question assesses the versatility of the world model to allow for new combinations of chains to be found by a planner that differ from any one recipe example in the training corpus.
That is, the world model supports creative trajectories through the state space and is not just memorizing what is necessary to recreate known plans. 

We maintain a holdout test set containing 200 never-before-seen actions and run automatic experiments to answer these two questions.
The answers are essentially determined by the diversity of our action precondition/effect corpus and the effectiveness of our precondition/effect inference modules.
Specifically, we iterate through all the actions in our corpus and use the effect inference module to infer their effects.
We also iterate through all the actions in the holdout test set and use the precondition inference module to infer their preconditions.

To answer the first question, for each action in the holdout test set, we use GPT-4o to compare each item in its precondition with each item in the corpus-level collection of all action effects, and to determine if there exists at least one effect item that semantically matches the precondition item.
If all the precondition items of this action are successfully matched with at least one effect item, then such an action is deemed satisfiable.
We find that 83.5\% of the actions in the holdout test set are satisfiable, suggesting that the search space created by our world model is large enough to be able to make most of the never-before-seen actions valid to be taken.
To answer the second question, for each action in the holdout test set, we use GPT-4o to compare each item in its precondition with each item in the corpus-level collection of all action effects, and to determine how many effect items there exist that semantically match the precondition item.
We multiply together all \# of matched effects for each precondition item and find that on average, a satisfiable action can be satisfied in 9.7 different ways in the search space.
It suggests that if one way to satisfy an action is blocked for some reason, one can reliably find other ways in this search space as alternatives.
It can also be inferred that the system is not simply memorizing dish recipes.

\section{Conclusions}

In this paper, we have demonstrated the potential of Large Language Models (LLMs) to function as world models by predicting valid actions and state transitions, two fundamental aspects of any world model.
Through fine-tuning, we adapt LLMs to infer both the preconditions and the effects of actions, thereby replicating key functions necessary for modeling environmental dynamics.
Our approach hinges on using synthetic data generation to enhance model training, a technique validated by human-participant studies, which confirms that the LLM-generated precondition and effect knowledge aligns with human understanding of real-world phenomena.
Automated evaluations further support the effectiveness and reliability of our method in creating a robust world model.

\section{Limitations}

While our approach demonstrates the ability to induce LLMs to behave as world models, it has certain limitations.
First, the model's performance is constrained by the quality and diversity of its training data, which may not cover super rare or highly novel scenarios.
Additionally, LLMs do not possess true causal reasoning, and their predictions rely on patterns from textual data rather than an inherent understanding of real-world dynamics.
Finally, the synthetic data generation process, while effective, may introduce biases or inaccuracies that could affect the overall performance in specific edge cases.

\bibliography{anthology,custom}

\appendix

\section{Implementation Details}
\label{sec:appendix-implementation-details}

We use Hugging Face Transformers\footnote{\url{https://github.com/huggingface/transformers}} \cite{wolf-etal-2020-transformers} during implementation.
We train the FLAN-T5-large models~\cite{chung2024scaling} using AdamW \cite{loshchilov2018decoupled} with the default learning rate linearly decaying from $5E-5$.
All models are trained for 50 epochs on an NVIDIA TITAN Xp GPU.
During inference, we perform greedy decoding.

\newpage
\onecolumn
\section{Prompts for Global-Local Prompting}
\label{sec:global-local}
\begin{small}
\begin{longtable}{p{0.9\textwidth}}

\toprule

Step \#1: (we denote the model output as \$model\_output\_step\_1) \\
\textbf{Give me a series of action steps that are generally involved in \$domain\_and\_task\_description.} \\
\textbf{Steps must have a strong dependency on each other.} \\
\textbf{Steps must be grounded in a specific concrete environment.} \\
\\
\textbf{A series of action steps:} \\

\midrule[0.03em]

Step \#2: (we denote the model output as \$model\_output\_step\_2) \\
\textbf{You will be given full action steps.} \\
\textbf{Identify and discard the action steps that you believe are isolated from and not quite dependent on other steps.} \\
\textbf{After discarding steps, if you find that without the discarded steps the full action steps lack coherence, you can optionally add new action steps in the place where you discard action steps, to make the full action steps look more coherent.} \\
\textbf{Unlike the discarded action steps, the new action steps you add must have a strong dependency on other existing steps.} \\
\\
\textbf{Below are some examples of how to identify and discard isolated and independent action steps.} \\
\textbf{\$few\_shot\_examples\_step\_2} \\
\\
\textbf{Full action steps:} \\
\textbf{\$model\_output\_step\_1} \\
\\
\textbf{Full action steps after discarding and adding steps:} \\

\midrule[0.03em]

Step \#3: (we denote the model output as \$model\_output\_step\_3) \\
\textbf{You will be given full action steps.} \\
\textbf{For each action step, independently infer all of its preconditions and effects which might comprise multiple sentences.} \\
\textbf{The precondition is the state that must be made true before action execution.} \\
\textbf{The effect is the state achieved after action execution.} \\
\textbf{Be as accurate and precise as possible.} \\
\textbf{Note that there is no need to make the precondition of step N identical to the effect of step N-1.} \\
\\
\textbf{Full action steps:} \\
\textbf{\$model\_output\_step\_2} \\
\\
\textbf{Full action steps with preconditions and effects:} \\

\midrule[0.03em]

Step \#4.1: (we denote the model output as \$model\_output\_step\_4.1) \\
\textbf{You will be given full action steps with preconditions and effects.} \\
\textbf{Identify the action steps (along with their preconditions and effects) that you believe have preconditions and effects which are isolated from and not quite dependent on other steps’ preconditions and effects.} \\
\\
\textbf{Below are some examples of how to identify isolated and independent action steps.} \\
\textbf{\$few\_shot\_examples\_step\_4.1} \\
\\
\textbf{Full action steps with preconditions and effects:} \\
\textbf{\$model\_output\_step\_3} \\
\\
\textbf{Identified action steps with preconditions and effects:} \\

\midrule[0.03em]

Step \#4.2: (we denote the model output as \$model\_output\_step\_4.2) \\
\textbf{You will be given full action steps with preconditions and effects.} \\
\textbf{You will also be given identified action steps with preconditions and effects that are isolated from and not quite dependent on other steps’ preconditions and effects.} \\
\textbf{Rethink and regenerate the preconditions and effects of these identified action steps, to make them more dependent on other steps’ preconditions and effects than before.} \\
\textbf{The precondition is the state that must be made true before action execution.} \\
\textbf{The effect is the state achieved after action execution.} \\
\textbf{Be as accurate and precise as possible.} \\
\\
\textbf{Full action steps with preconditions and effects:} \\
\textbf{\$model\_output\_step\_3} \\
\\
\textbf{Identified action steps with preconditions and effects:} \\
\textbf{\$model\_output\_step\_4.1} \\
\\
\textbf{Identified action steps with newly generated preconditions and effects:} \\

\midrule[0.03em]

Step \#4.3: (we denote the model output as \$model\_output\_step\_4.3) \\
\textbf{You will be given full action steps with preconditions and effects.} \\
\textbf{You will also be given identified action steps with preconditions and effects that are isolated from and not quite dependent on other steps’ preconditions and effects.} \\
\textbf{You will also be given identified action steps with newly generated preconditions and effects.} \\
\textbf{Categorize these identified action steps (with newly generated preconditions and effects) into the following two groups: (1) identified action steps whose newly generated preconditions and effects look dependent on other steps’ preconditions and effects; (2) identified action steps whose newly generated preconditions and effects look not quite dependent on other steps’ preconditions and effects.} \\
\\
\textbf{Below are some examples of how to do this categorization.} \\
\textbf{\$few\_shot\_examples\_step\_4.3} \\
\\
\textbf{Full action steps with preconditions and effects:} \\
\textbf{\$model\_output\_step\_3} \\
\\
\textbf{Identified action steps with preconditions and effects:} \\
\textbf{\$model\_output\_step\_4.1} \\
\\
\textbf{Identified action steps with newly generated preconditions and effects:} \\
\textbf{\$model\_output\_step\_4.2} \\
\\
\textbf{Categorization:} \\

\midrule[0.03em]

Step \#5.1: \\
\textbf{You will be given full action steps with preconditions and effects.} \\
\textbf{Find preconditions that are semantically not covered in any of the effects.} \\
\textbf{Semantic coverage means there exists at least one effect item that expresses the semantically equivalent meaning as the precondition item.} \\
\\
\textbf{Full action steps with preconditions and effects:} \\
\textbf{\$model\_output\_step\_4.3\_after\_post\_processing} \\
\\
\textbf{Preconditions that are semantically not covered in any of the effects:} \\

\midrule[0.03em]

Step \#5.2: \\
\textbf{You will be given full action steps with preconditions and effects.} \\
\textbf{Find effects that are semantically not covered in any of the preconditions.} \\
\textbf{Semantic coverage means there exists at least one precondition item that expresses the semantically equivalent meaning as the effect item.} \\
\\
\textbf{Full action steps with preconditions and effects:} \\
\textbf{\$model\_output\_step\_4.3\_after\_post\_processing} \\
\\
\textbf{Effects that are semantically not covered in any of the preconditions:} \\

\bottomrule
\caption{The prompts we use in global-local prompting (\autoref{subsec:module1}). We omit the output-format controlling prompts for brevity.}
\label{tab:global-local}
\end{longtable}
\end{small}

\newpage
\onecolumn
\section{Prompts for Semantic State Matching}
\label{sec:semantic-matching}
\begin{small}
\begin{longtable}{p{0.9\textwidth}}

\toprule

Prompts used in the semantic state matching for valid action prediction: \\
\textbf{You will be given some precondition items.} \\
\textbf{You will also be given some world-state items.} \\
\textbf{For each precondition item, determine if it is semantically covered by the world-state items.} \\
\textbf{Semantic coverage means there exists at least one world-state item that expresses the semantically equivalent meaning as the precondition item.} \\
\textbf{If all the precondition items are semantically covered, return TRUE; otherwise, return FALSE.} \\
\\
\textbf{Precondition items:} \\
\textbf{\$inferred\_action\_preconditions} \\
\\
\textbf{World-state items:} \\
\textbf{\$current\_world\_state} \\

\midrule[0.03em]

Prompts used in the semantic state matching for state transition prediction: \\
\textbf{You will be given some effect items.} \\
\textbf{You will also be given some world-state items.} \\
\textbf{For each effect item, find all the world-state items that semantically contradict the effect item.} \\
\\
\textbf{Effect items:} \\
\textbf{\$inferred\_action\_effects} \\
\\
\textbf{World-state items:} \\
\textbf{\$current\_world\_state} \\

\bottomrule
\caption{The prompts we use in semantic state matching (\autoref{subsec:module2}). We omit the output-format controlling prompts for brevity.}
\label{tab:semantic-matching}
\end{longtable}
\end{small}

\end{document}